\documentclass{article}
\usepackage{spconf,amsmath,graphicx,hyperref}

\usepackage{comment}
\usepackage{multirow}
\usepackage{colortbl}
\usepackage{xcolor}
\usepackage{subcaption}
\usepackage{booktabs} % Required for the \toprule, \midrule, \bottomrule commands
\usepackage{caption}  % Required for the caption styling

\title{Evaluating Emotion Recognition in Spoken Language Models on Emotionally Incongruent Speech}
\name{Pedro Corrêa, João Lima, Victor Moreno, Lucas Ueda, Paula Costa}

\address{School of Electrical and Computer Engineering\\
	Universidade Estadual de Campinas (UNICAMP)\\
	Campinas, Brazil}
\begin{document}
%\ninept
%
\maketitle
\begin{abstract}

% Speech-aware spoken language models (SLMs) are a subset of models within the larger field of spoken language models (SLMs). Despite methodological differences, these models share the same general goal: to extract and interpret information from speech in a text-instructed way in order to complete a given task, the result also being in text format. While they show strong performance on speech and emotion recognition, we found that SLMs exhibit bias when tested on emotionally incongruent speech (when the emotion present in the audio content differs from the present in the audio tone). In this setting, all evaluated SLMs favored textual content over acoustic cues, such as emotional tone, when presented with synthetic speech. Objective metrics reveal substantial performance gaps between emotion labels derived from text and those from audio. A user study was conducted to verify the emotional consistency of synthetic speech audios and their labeled emotion in this evaluation.

Advancements in spoken language processing have driven the development of spoken language models (SLMs), designed to achieve universal audio understanding by jointly learning text and audio representations for a wide range of tasks. Although promising results have been achieved, there is growing discussion regarding these models' generalization capabilities and the extent to which they truly integrate audio and text modalities in their internal representations. In this work, we evaluate four SLMs on the task of speech emotion recognition using a dataset of emotionally incongruent speech samples, a condition under which the semantic content of the spoken utterance conveys one emotion while speech expressiveness conveys another. Our results indicate that SLMs rely predominantly on textual semantics rather than speech emotion to perform the task, indicating that text-related representations largely dominate over acoustic representations. We release both the code and the Emotionally Incongruent Synthetic Speech dataset (EMIS) to the community.

% PODE TER ATÉ NO MÁXIMO 150 PALAVRAS!!!
% TÁ COM EXATAMENTE 150. SE ALTERAR, CHECAR NOVAMENTE.

\end{abstract}
\section{Introduction}
\label{sec:intro}
% Spoken language technologies increasingly rely on SLMs that jointly process acoustic signals and linguistic content. Recent SLMs extend this trend, promising to go beyond conventional automatic speech recognition (ASR) by leveraging prosody, timbre, and other paralinguistic cues for tasks such as dialog understanding, affective computing, and multimodal grounding. However, despite rapid progress, it remains unclear whether current SLMs genuinely interpret acoustic–prosodic information or whether their predictions are dominated by semantic priors extracted from lexical content. %

Spoken language technologies are increasingly relying on spoken language models (SLMs) that combine acoustic and semantic information within a unified framework~\cite{arora2025landscape}. Unlike conventional pipelines that integrate automatic speech recognition, large language models (LLMs), and text-to-speech (TTS) modules, recent SLMs aim for end-to-end modeling. They capture not only semantic content, but also prosody, timbre, and other paralinguistic cues essential for a better world understanding.
%[... aim for end-to-end modeling.] They capture not only semantic content, but also prosody, timbre, and other paralinguistic cues essential for a better world understanding.
This path mirrors the evolution of text-based natural language processing, which advanced from task-specific models to universal LLMs. However, SLMs last at an earlier stage. Current approaches are categorized as pure speech models trained in tokenized audio, joint speech–text models that exploit paired data, and speech-aware SLMs that combine speech encoders with pretrained LLMs~\cite{arora2025landscape}. The latter (henceforth referred to as SLMs) are the focus of this work. These models receive speech and text as input and, as output, an answer in text format by using the instruction-following capabilities of LLMs. Despite progress, it is unclear whether SLMs actually retain information from acoustic-prosodic signals or default to semantic information, highlighting the need for systematic investigation of their decision processes~\cite{chi2025roleofprosody}.

Emotion recognition provides a probe to address this question, as semantic and prosodic channels are not always aligned~\cite{kikutani2022detecting}. Most SLM evaluations focus on congruent examples, where both channels convey the same emotion (e.g., ``\textit{I am sad}'' spoken in a sad tone)~\cite{busso2008iemocap}. In such settings, models may detect explicit or implied emotion from words alone, bypassing paralinguistic cues such as pitch, intensity, and rhythm. In contrast, incongruent cases, where semantic content and prosody conflict (e.g., ``\textit{I am sad}'' but spoken happily despite conveying a negative sentiment), are rarely evaluated. Previous studies show that prosody and semantic content can exert competing influences under incongruence, reinforcing the need for benchmarks that separate these channels~\cite{kikutani2022detecting}.

%We address this gap by designing a controlled
We address this gap by designing a controlled evaluation in which semantic and prosodic cues can be explicitly aligned or placed in conflict. Synthetic speech samples, both congruent and incongruent, are generated with state-of-the-art (SoTA) TTS systems conditioned on emotional reference recordings. These samples cover the case when the emotion tag is stated directly in the utterance, when sentiment is implied through context, and when it is neutral. This setup enables the disentanglement of acoustic and semantic contributions in SLM decisions by testing whether their predictions are based on speech expressiveness or on semantic content.

%reveal that SLMs rely predominantly on textual semantics rather than speech emotion to perform the task

%Through controlled experiments, we demonstrate that prosodic information alone can, to some extent, guide models to answer questions in SQA tasks. However, while prosody offers meaningful complementary cues, we find that models predominantly rely on lexical information when it is available.

Our contributions are (i) the observation that evaluated SLMs rely predominantly on semantic content rather than speech expressiveness to perform emotion recognition, using an evaluation protocol that contrasts SLMs with a baseline acoustic speech emotion recognition system (SER) and human listeners, and (ii) the creation of the Emotionally Incongruent Synthetic Speech dataset (EMIS)\footnote{\href{https://ieee-dataport.org/documents/emotionally-incongruent-synthetic-speech-dataset-emis}{Emotionally Incongruent Synthetic Speech dataset (EMIS)}}. Code in Github\footnote{\href{https://github.com/AI-Unicamp/SLM-ER-Evaluation}{Github Repository}}.

% elaboration of an evaluation protocol that contrasts SLMs with a baseline acoustic speech emotion recognition (SER) system and human listeners, revealing systematic semantic bias in current SLMs and (ii) the creation of the Emotionally Incongruent Synthetic Speech dataset (EMIS) \footnote{\href{https://ieee-dataport.org/documents/emotionally-incongruent-synthetic-speech-dataset-emis}{Emotionally Incongruent Synthetic Speech dataset (EMIS)}}. Code is available at Github\footnote{\href{https://github.com/AI-Unicamp/SLM-ER-Evaluation}{Github Repository}}.

%Our results reveal a systematic semantic bias, in which SLMs tend to align their predictions with the lexical content of the utterance rather than with the emotion conveyed in the vocal tone, a tendency particularly pronounced when the text explicitly names an emotion. In contrast, the baseline SER and human listeners consistently followed the prosodic signal, confirming that it is both salient and perceptible. These findings suggest that current SLMs, even when prompted to rely exclusively on vocal tone, remain dominated by semantic cues rather than prosodic information. To support this conclusion, we present a dataset of emotionally incongruent synthetic speech, an evaluation protocol that compares SLMs with an acoustic SER baseline and human listeners under prosody-only prompting, and empirical evidence that exposes the semantic bias underlying current models.

\section{Related Works}
\label{subsec:generating}

SLMs extend instruction-following LLMs to operate on speech by mapping audio into compact representations interpretable by the language model. Recent systems are trained on multiple tasks, including emotion recognition, and differ in scope and training strategy: SALMONN~\cite{tang2024salmonn} targets speech, general audio, and music, arguing that joint training across heterogeneous audio domains yields broad capabilities and introducing techniques to preserve emergent abilities after instruction tuning; DeSTA2~\cite{lu2025desta2} forgoes speech instruction-tuning by supervising with automatically generated, domain-agnostic speech captions, aiming to retain the base LLM’s reasoning; Qwen2-Audio~\cite{chu2024qwen2audio} follows a three-stage alignment pipeline to strengthen instruction following and user-aligned behavior over audio inputs; and Audio Flamingo 3~\cite{goel2025audioflamingo} emphasizes general-audio use cases with long-context interaction, multi-audio dialogue, and chain-of-thought prompting, trained via a multi-stage curriculum on open data.

% In practice, systems are trained and evaluated on multitask settings, covering ASR, captioning/QA, and paralinguistic objectives such as speaker emotion recognition across diverse corpora.

In contrast, acoustic SER systems estimate emotion from the signal alone, relying on prosodic evidence~\cite{ma2023emotion2vecselfsupervisedpretrainingspeech}. This makes SER a prosody-centric reference for interpreting the behavior of SLM under semantic-prosodic incongruence. Fair analysis requires decoupling semantic content and speech expressiveness during evaluation.
%In our study, we compare SLMs with an acoustic SER baseline and human perceptual judgments to assess whether model decisions follow the voice or the words.
Chi et al.~\cite{chi2025roleofprosody} propose isolating prosodic and semantic information in spoken question-answer by low-pass filtering the audio signal (prosody) and by flattening pitch and intensity (lexical), finding that models perform reasonably well with prosody alone, but predominantly rely on semantic cues when text is present. Furthermore, Kikutani~\cite{kikutani2022detecting} analyzes human judgments of speech expressing incongruent emotional cues through voice and content, revealing that cue dominance varies across languages and modalities. 
We therefore bring the test to emotion recognition, but instead of removing the semantic content of the signal, we induce a controlled semantic-prosodic incongruence.

\section{Methods}
\label{sec:method}

\begin{figure}[htbp]
    \centering
    \includegraphics[width=1.0\linewidth]{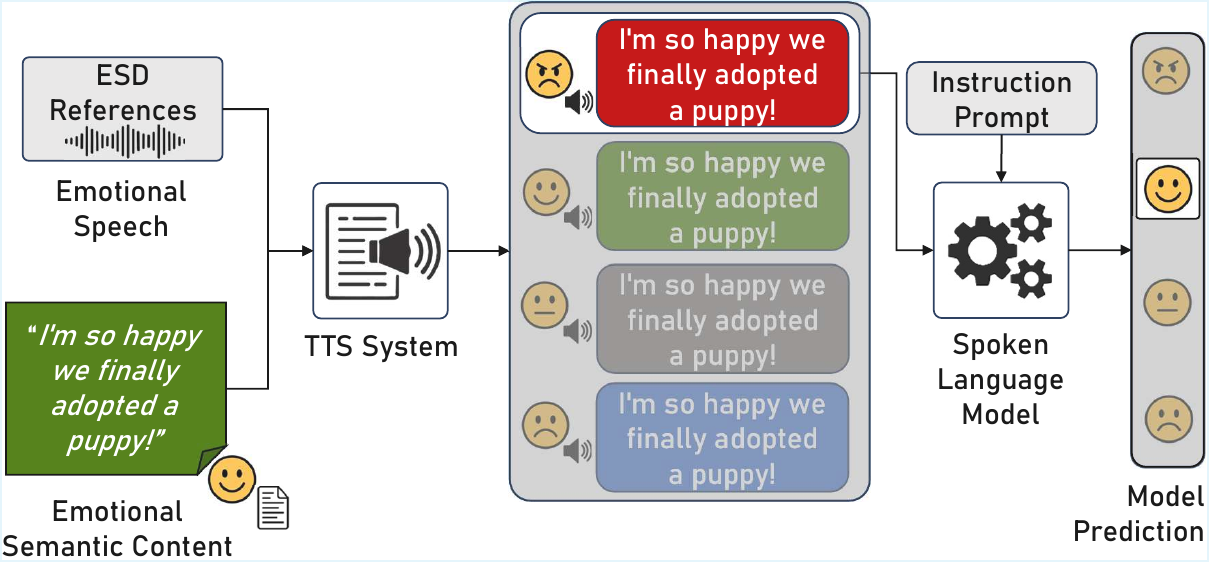}
    \caption{An emotion-rich sentence is paired with emotional speech references from the ESD dataset to form the TTS input. For each text, the TTS system generates four speech samples, one for each acoustic expressiveness (angry, happy, neutral, sad), derived from the ESD audio references. Each generated sample is then analyzed with an SLM guided by an instruction prompt to classify the conveyed emotion.}
    \label{fig:general_pipeline}
\end{figure}

Our proposed evaluation protocol (Figure \ref{fig:general_pipeline}) consists of first generating emotion-rich sentences using an LLM, then generating synthetic speech samples by providing TTS systems with these sentences alongside emotional reference speech. We assess the quality of the synthetic speech by employing a baseline SER model and conducting a human perceptual evaluation to verify if the reference emotions are correctly identified in each generated sample. Finally, we prompt the SLMs to perform the emotion recognition task on the generated speech samples, extract, and analyze the results.

\subsection{Generating Speech Samples}
\label{subsec:generating}
% \begin{itemize}
%     \item geração dos textos a partir do chatgpt4.5 com controle do sentimento da frase.
%     \item gera-se 26 frases para cada emoção (13/13 exp./imp.) e 26 neutras
%     \item utilização de sentimento implicito e explicito nas frases
%     \item cada frase passa por um dos tres TTS utilizados. cada frase gera 4 samples, cada um em uma emoção (neutral, sad, happy, angry)
%     \item para a geração dos samples, a emoção vem de audios de referencia retirados dos falantes em ingles do dataset ESD (subset que foi test set do SER)
%     \item cada TTS gera 416 samples
% \end{itemize}

% \subsubsection{Emotion-rich Sentences}
% \label{subsubsec:sentences}

We employ GPT-4.5 to generate 104 emotion-rich sentences divided into 4 distinct emotions: angry, happy, neutral, and sad.
%\footnote{\href{https://ieee-dataport.org/documents/emotionally-incongruent-synthetic-speech-dataset-emis}{All used sentences are provided alongside EMIS}}.
Emotion-rich sentences can be defined as natural language text that refers the reader back to a sentiment. We divide these emotion classes into two categories each: explicit and implicit. Explicit samples contain the exact emotion tag (e.g., ``\textit{I'm so happy we finally adopted a puppy!}''), whereas implicit samples do not contain the tag, conveying emotion from the context (e.g., ``\textit{I can’t stop smiling after our date last night.}''). This distinction doesn't apply to neutral sentences since there's no conveyed emotion; thus, we analyze them as a separate condition in our experiments.
% \subsubsection{Emotionally Incongruent Speech}
% \label{subsubsec:sentences}

Recent advances in zero-shot expressive TTS have made it possible to synthesize speech with controllable emotional styles by extracting expressiveness from short reference recordings and transferring it to the generated output. Beyond improving naturalness, these systems allow the generation of stimuli where semantic content and prosodic realization can be independently manipulated, creating congruent and incongruent pairs at scale. This capability enables the experimental foundation of this work.

%For each of the 104 emotion-rich sentences, we employ three TTS systems to generate four speech samples, one for each of the four emotions used in sentence generation. Thus, we end up with three emotionally incongruent speech samples for each emotion-rich sentence. The use of three different TTS systems (CosyVoice2.0, StyleTTS-2, and F5-TTS) aims to mitigate potential biases from each system training stage~\cite{du2024cosyvoice2scalablestreaming, li2023styletts2, chen2025f5ttsfairytalerfakesfluent}.

We employ three distinct SoTA TTS models~\cite{du2024cosyvoice2scalablestreaming, li2023styletts2, chen2025f5ttsfairytalerfakesfluent} to generate four speech samples for each of the 104 emotion-rich sentences, each sample corresponding to one of the four emotions. Thus, the resulting dataset (EMIS) contains one emotionally congruent and three incongruent speech samples for each emotion-rich sentence. In the incongruent condition, we treat the emotion conveyed by the speech signal as the target label for the emotion recognition task and the one conveyed by the semantic content as the proxy label.  

Since all three systems require reference audios for inference, we extract them from English speakers in The Emotional Speech Database (ESD)~\cite{zhou2021seenandunseen}. ESD comprises 10 English speakers with 350 utterances per emotion. To extract reference samples, we randomly select a speaker, then select and concatenate their 7 longest utterances to create longer reference audios with approximately 30 seconds (mean and standard deviation are 32.2 and 3.5 seconds, respectively). Each TTS generates 416 samples (Angry, Happy, Neutral, and Sad emotions for each of the 104 emotion-rich sentences), resulting in the EMIS dataset with a total of 1248 speech samples.

%\subsection{Evaluation of Spoken Language Models on ER Task}
\label{subsec:generating}
% \begin{itemize}
%     \item utilização de 4 SLMs na task de emotion recogtion
%     \item o mesmo prompt ``\textit{Using tone of voice only (prosody: pitch, rhythm, loudness, timbre). Ignore word meaning; do not transcribe. Reply with exactly one: angry | happy | sad | neutral}.'' para todos os modelos
%     \item Obtenção de métricas objetivas para comparações
% \end{itemize}

% We employ four SLMs (DeSTA2, Audio Flamingo-3, Qwen-2-Audio, and SALMONN) to be evaluated on the task of emotion recognition with our synthetic speech samples \cite{lu2025desta2, goel2025audioflamingo, chu2024qwen2audio, tang2024salmonn}. We constructed a single text prompt to instruct all models in the task: ``\textit{Using tone of voice only (prosody: pitch, rhythm, loudness, timbre). Ignore word meaning; do not transcribe. Reply with exactly one: angry | happy | sad | neutral}''. This prompt was constructed to avoid the model to generate different emotions other than the mentioned four, as well as instructing the model to extract information solely from the tone of the voice, and not the transcribed text.

% \section{Experiments}
% \label{sec:experiments}

% \begin{itemize}
%     \item tabela de comparação com métricas objetivas entre modelos, TTS e audio/text label (Table \ref{tab:model_eval})
%     \item tabelas de comparação (para cada modelo) com métricas objetivas entre emoções do texto e o audio (Tables \ref{tab:emotion_eval_desta2}, \ref{tab:emotion_eval_flamingo3}, \ref{tab:emotion_eval_qwen2audio}, \ref{tab:emotion_eval_salmonn})
%     \item 
% \end{itemize}

\begin{table*}[ht]
\centering
\caption{Target (audio emotion) and proxy (semantic content emotion) accuracy scores achieved by each SLM, as well as the baseline SER system, under each semantic category defined in \ref{subsec:generating}. SLMs' target accuracies are consistently low across conditions, whereas the modality-specific baseline SER exhibits superior performance. SLMs predict proxy emotions more consistently in the \textit{explicit} semantic condition and show distinct patterns of behavior in \textit{implicit} and \textit{neutral} conditions.}
\label{tab:acc_txtCond}
\begin{tabular}{@{}llcccccc@{}}
\toprule
\multicolumn{2}{c}{} & \multicolumn{2}{c}{\textbf{Explicit Category}} & \multicolumn{2}{c}{\textbf{Implicit Category}} & \multicolumn{2}{c}{\textbf{Neutral Category}} \\
\cmidrule(lr){3-4} \cmidrule(lr){5-6} \cmidrule(lr){7-8}
 \textbf{SLM} & \textbf{TTS} & \textbf{Acc. (\%)} & \textbf{Proxy Acc. (\%)} & \textbf{Acc. (\%)} & \textbf{Proxy Acc. (\%)} & \textbf{Acc. (\%)} & \textbf{Proxy Acc. (\%)} \\
\midrule
\multirow{3}{*}{DeSTA2} & CosyVoice2 & 25.6 & 95.5 & 30.1 & 89.1 & 34.6 & 8.6 \\
\multirow{3}{*}{} & F5-TTS & 25.6 & 95.5 & 25.0 & 89.7 & 29.8 & 10.5 \\
\multirow{3}{*}{} & StyleTTS2 & 25.6 & 97.4 & 28.2 & 91.6 & 38.4 & 7.6 \\
\midrule
\multirow{3}{*}{Audio Flamingo3} & CosyVoice2 & 28.8 & 93.5 & 37.8 & 66.0 & 41.3 & 76.9 \\
\multirow{3}{*}{} & F5-TTS & 26.2 & 98.7 & 31.4 & 82.6 & 38.4 & 86.5 \\
\multirow{3}{*}{} & StyleTTS2 & 25.0 & 100.0 & 30.1 & 82.0 & 37.5 & 82.6 \\
\midrule
\multirow{3}{*}{Qwen2Audio} & CosyVoice2 & 26.2 & 96.7 & 30.1 & 69.2 & 21.1 & 11.5 \\
\multirow{3}{*}{} & F5-TTS & 26.2 & 98.7 & 29.4 & 75.6 & 26.9 & 9.6 \\
\multirow{3}{*}{} & StyleTTS2 & 25.6 & 99.3 & 29.4 & 73.0 & 26.9 & 6.7 \\
\midrule
\multirow{3}{*}{SALMONN} & CosyVoice2 & 28.9 & 80.2 & 25.6 & 21.1 & 25.9 & 89.4 \\
\multirow{3}{*}{} & F5-TTS & 26.9 & 80.9 & 33.3 & 23.7 & 26.9 & 92.3 \\
\multirow{3}{*}{} & StyleTTS2 & 27.2 & 89.6 & 26.2 & 30.1 & 36.5 & 71.1 \\
\midrule
\multirow{3}{*}{\textbf{Baseline SER}} & CosyVoice2 & 52.5 & 31.4 & 53.2 & 33.3 & 47.1 & 9.0 \\
\multirow{3}{*}{} & F5-TTS & 48.0 & 31.4 & 46.1 & 33.3 & 50.0 & 8.6 \\
\multirow{3}{*}{} & StyleTTS2 & 50.0 & 26.9 & 47.4 & 30.7 & 49.0 & 1.0 \\
\bottomrule
\end{tabular}
\end{table*}

\subsection{Experimental Setup}
% \subsubsection{Text Prompt Construction}

We employ four SLMs (Audio Flamingo-3, DeSTA2, Qwen-2-Audio, and SALMONN) to be evaluated on the task of emotion recognition with synthetic speech samples~\cite{lu2025desta2, goel2025audioflamingo, chu2024qwen2audio, tang2024salmonn}. Since SLMs have LLMs as their backbones, these models are very prompt-sensitive. For this reason, we carefully build a single text prompt to instruct all models in the task: ``\textit{Using tone of voice only (prosody: pitch, rhythm, loudness, timbre). Ignore word meaning; do not transcribe. Reply with exactly one: angry | happy | sad | neutral}''. This prompt was constructed to guide the model to avoid generating different emotions other than the chosen four, as well as to instruct the model to extract information solely from the acoustic expressiveness of the voice, and not the semantic content. During inference, we relied on each spoken language model’s default hyperparameter configuration, due to the values being similar across models, and to avoid altering settings optimized during their development.

% We have tested three other prompts before choosing the final one. Since SLMs have a LLM in their backbone, these models are very prompt-sensitive. For this reason, we carefully build the prompt, testing simpler and more complex ones. 

% While simpler ones lead to the models generating answers out of the four expected emotions, we needed a more complex prompt that would limit this aspect. However, with longer and even more complex prompts, older models such as SALMONN would struggle to understand the task and end up hallucinating.

% \subsubsection{Models' Hyperparameters}

% The evaluations in this work are based on outputs generated by the large language model (LLM) end of SLMs, which are inherently non-deterministic. To ensure consistency, we used the same text prompts across all models and carefully considered the values of the Temperature and Nucleus Sampling (TOP-P) parameters. Although the models are instructed to generate only a single word, these parameters still exert a small but non-negligible influence on the outputs. For inference, we relied on each spoken language model’s default hyperparameter configuration, due to the values being similar across models, and to avoid altering settings optimized during their development.

\subsection{Evaluation Metrics}

% \subsubsection{Objective Metrics}

To objectively measure our evaluation results, we employ metrics to assess model bias towards semantic information on the emotion recognition task. Each SLM sequentially receives as input all samples from the EMIS dataset alongside the textual instruction prompt. We compare the models' outputs with respect to the investigated conditions (congruency and semantic emotion explicitness) by analyzing accuracy scores relative to the target and proxy labels and performing statistical chi-square hypothesis tests.  

%[... based on the] emotion label for both the text and audio. Leveraging objective metrics such as accuracy and chi-squared, we are able to analyze the results of each SLM.

In addition, we also finetune a Speech Emotion Recognition (SER) model on a subset of the ESD dataset to validate the quality of the TTS-generated synthetic speech samples \cite{ma2023emotion2vecselfsupervisedpretrainingspeech}. Since we use these samples to conduct our main evaluation, we first verify their reliability regarding the emotion conveyed by speech expressiveness.

We conducted two chi-squared tests of independence to investigate whether the distribution of model predictions depends on the target and proxy labels. The tests compare observed frequencies of model predictions against frequencies expected under statistical independence. For each analysis, we constructed contingency tables from 4,978 samples, with nine degrees of freedom, given the four emotion classes. The null hypothesis stated that no significant association exists between the observed variables, while the alternative hypothesis posited a significant association. We also calculate the effect size for each test using Cramér's V statistic.

% \subsubsection{Subjective Metrics}

To further assess the reliability of the generated speech samples, we conduct a human perceptual evaluation as an additional validation step. By asking participants to identify emotion conveyed in synthetic speech expressiveness, we can verify consistency between the labeled reference emotions and those detected by humans. The perceptual evaluation was conducted with 40 participants on a balanced subset of the EMIS dataset. The results of users' accuracy are divided between TTS systems and ground-truth samples. They are summarized as: $39.4\%$ for StyleTTS2, $58.1\%$ for CosyVoice2, $62.0\%$ for F5-TTS, and $70.8\%$ for ground-truth.

%This subset is balanced across (1) TTS systems, with 16 samples per model; (2) speech emotions, with 4 samples per emotion; (3) speakers, with approximately 6 samples per speaker. Moreover, we finish populating this subset with ground-truth examples extracted directly from the ESD dataset, while keeping it balanced. In total, we have a subset with 64 samples.

\section{Results and Discussion}

% Results from the human perceptual evaluation are summarized in Table~\ref{tab:user_study}. Participants achieved high accuracy scores when classifying speech emotion in samples generated by CosyVoice2 and F5-TTS (58.1\% and 62\%, respectively), suggesting that these systems effectively capture speech emotion from reference samples to generate the intended audio, making them suitable for our subsequent experiments. StyleTTS2 performed worse, which we hypothesize may be due to the reference audio concatenation procedure, potentially introducing unnatural utterance flow that affects generation quality. Nevertheless, participants' accuracy on StyleTTS2 samples remains non-negligible for a four-class classification problem, so we include this system in the following experiments.      

% \begin{table}[htbp]
%     \centering
%     \caption{User Study Results of Accuracy of Participants on Identifying the Labeled Emotion of Speech Samples Generated by Each Source}
%     \begin{tabular}{c|c|c|c}
%     \hline
%     \textbf{StyleTTS2} & \textbf{CosyVoice2} & \textbf{F5-TTS} & \textbf{Ground-Truth} \\
%     \hline
%     39.4\% & 58.1\% & 62.0\% & 70.8\% \\
%     \hline
%     \end{tabular}
    
%     \label{tab:user_study}
% \end{table}

Once the experimental setup was validated, we proceeded to evaluate the SLMs. Table~\ref{tab:acc_txtCond} shows accuracy scores achieved by each SLM for predicting both the target (audio) and proxy (semantic content) emotion labels. For comparison, the performance of the baseline SER system is also reported. Accuracy scores relative to target audio emotions approach those of a random classifier (25\% for a four-class setting), whereas those relative to proxy labels are considerably higher under most conditions. There are substantial gaps between SLMs' target and proxy accuracies across all semantic categories, most pronounced in the explicit case, in which Audio Flamingo3 notably displays a categorical pattern, always predicting the proxy label when classifying StyleTTS2 samples. 

For the neutral category, accuracies remained stable for Qwen2Audio and SALMONN, but improved for DeSTA2 and Flamingo3 when compared with explicit and implicit categories. These results indicate that in the absence of emotional cues from semantic content, some SLMs appear to more effectively leverage acoustic information to perform emotion recognition. Moreover, this category led DeSTA2 and Qwen2Audio to perform significantly worse in proxy accuracy, whereas SALMONN performed slightly better, which can be associated with the text sentiment analysis capabilities of each model. In contrast, the modality-specific baseline SER consistently achieves higher target and lower proxy accuracies, indicating a focus on prosody cues, the desired behavior for this validation model. These results support the argument that SLMs have a strong tendency to prioritize information present in the semantic content rather than speech acoustics to perform the task, especially when the semantic content is not neutral.

\begin{figure}[htbp] 
    \centering

    % First subfigure (a)
    \begin{subfigure}{0.49\columnwidth}
        \centering
        \caption{Congruent Condition}
        \includegraphics[width=\linewidth]{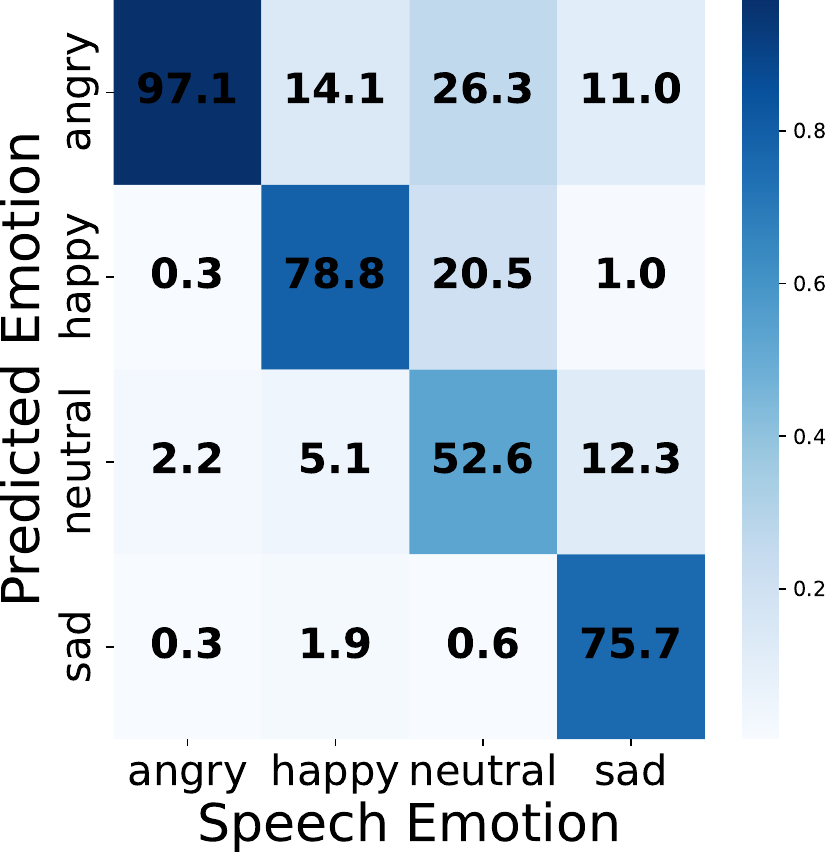}
        \label{fig:congruent}
    \end{subfigure}
    \hfill
    % Second subfigure (b)
    \begin{subfigure}{0.49\columnwidth}
        \centering
        \caption{Incongruent Condition}
        \includegraphics[width=\linewidth]{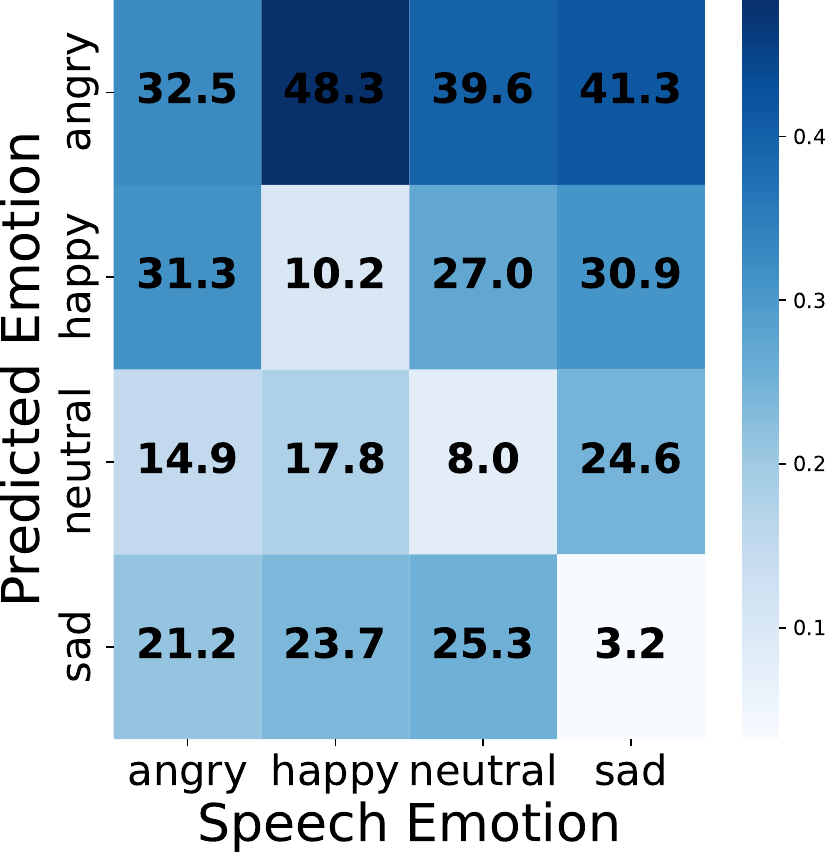}
        \label{fig:incongruent}
    \end{subfigure}

    \caption{Speech Emotion (target) vs. Predicted Emotion under congruent (a) and incongruent (b) conditions. Prediction counts are normalized column-wise and presented as percentage values. SLM predictions closely match the target labels when evaluated only on congruent samples, but display irregular, less reliable behavior with incongruent samples.}
    \label{fig:emotion_heatmaps}
\end{figure}

Class-specific SLM decisions are presented in Figure \ref{fig:emotion_heatmaps}. Under the congruent condition, i.e., when speech and semantic content have matching emotions, target and predicted emotions are closely aligned, indicating that SLM systems apparently leverage information mutually present in speech and semantic content. However, under the incongruent condition, i.e., when speech emotion differs from semantic content, this alignment breaks down, and SLM systems exhibit clear tendencies towards predicting the \textit{angry} and \textit{happy} classes while overlooking the \textit{sad} class. This may reflect an interaction effect between the way each SLM model captures information and the fact that \textit{angry} and \textit{happy} samples are more closely associated with prosodic variations than \textit{sad} and \textit{neutral}.

The conducted chi-squared tests indicated that predicted emotion is significantly associated with both target and proxy labels ($p < 0.01$ for both cases), allowing us to reject the null hypothesis. However, the association between predicted and target emotions exhibited a very small effect size, with a Cramér's $V$ of 0.08, whereas the association between predicted and proxy emotions showed a considerable effect size ($V = 0.65$). These findings suggest that while acoustic cues have some influence on the models' decisions, they are largely overshadowed by the spoken utterances' semantic content, which has a much stronger impact on the model's prediction.

\section{Conclusion}
\label{sec:conclusion}

%This work investigated the influence of semantic content present in spoken utterances on the speech understanding capabilities of speech-aware Spoken Language Models. We propose an evaluation of speech emotion recognition capabilities by employing emotionally incongruent speech, a condition under which the spoken content conveys one emotion, and speech expressiveness conveys a distinct one. 

This work investigated whether current SLMs can truly integrate semantic and acoustic information in their internal representations. Although seen as steps toward universal audio understanding, our evaluation suggests that these models fall short of this goal, showing a limited ability to disentangle semantics and acoustics when conflicting. The obtained results show that there is an imbalance between text and audio modalities, as the models tend to over-rely on information present in textual semantics, the more easily available that information is, as we can see in the explicit semantic condition. This has major implications for the rapidly growing ecosystem of speech foundational models. If these models are evaluated primarily on benchmarks where semantic content and acoustic expressiveness are aligned, their apparent competence may mask critical deficiencies in their capacity for paralinguistic reasoning, crucial component in applications that depend on nuanced interpretation of human communication, such as detecting irony, sarcasm, or emotional subtleties.

% The obtained results show that SLMs tend to be heavily influenced by semantic content. The experiments show that there is an imbalance between text and audio modalities, and the models tend to over-rely on information present in the speech transcripts, the more easily-available that information is, as we can see in the explicit semantic condition. This has major implications for the rapidly growing ecosystem of speech foundational models. If these models are evaluated primarily on benchmarks where semantic content and acoustic expressiveness are aligned, their apparent competence may mask critical deficiencies in their capacity for paralinguistic reasoning.

% Future work is needed to further explore the reasons why this modality imbalance emerges, as well as to propose mitigation strategies to achieve more robust multimodal integration. Architectures that more tightly couple speech representations with higher-level reasoning may be necessary to achieve true general audio understanding. Equally important, evaluation benchmarks must deliberately decouple semantic and acoustic signals to expose the degree to which models genuinely exploit prosodic and paralinguistic information.

% \clearpage
\section{Acknowledgment}
This work was partially funded by the Coordenação de Aperfei\c{c}oamento de Pessoal de Nível Superior – Brasil (CAPES) – Finance Code 001, by the S\~{a}o Paulo Research Foundation (FAPESP) under grant \#2020/09838-0 (BI0S - Brazilian Institute of Data Science) and \#2023/12865-8 (Horus project), and by the Ministry of Science, Technology, and Innovations, with resources from Law No. 8.248, of October 23, 1991, under the PPI-SOFTEX program, DOU 01245.003479/2024-10. The authors are also affiliated with the Artificial Intelligence Lab, Recod.ai.

\section{Compliance with Ethical Standards statement}
\label{sec:Compliance}

Part of this research study was conducted retrospectively using human subject data extracted from the employed human perceptual evaluation. This evaluation has been approved by the Ethical Committee on Research (CEP) from Universidade Estadual de Campinas (UNICAMP) under CAAE Number 59536022.8.0000.5404.

% \section{Acknowledgment}
% \label{sec:conclusion}

% To achieve the best rendering both in printed proceedings and electronic proceedings, we
% strongly encourage you to use Times-Roman font.  In addition, this will give
% the proceedings a more uniform look.  Use a font that is no smaller than nine
% point type throughout the paper, including figure captions.

% In nine point type font, capital letters are 2 mm high.  {\bf If you use the
% smallest point size, there should be no more than 3.2 lines/cm (8 lines/inch)
% vertically.}  This is a minimum spacing; 2.75 lines/cm (7 lines/inch) will make
% the paper much more readable.  Larger type sizes require correspondingly larger
% vertical spacing.  Please do not double-space your paper. TrueType or
% Postscript Type 1 fonts are preferred.

% The first paragraph in each section should not be indented, but all the
% following paragraphs within the section should be indented as these paragraphs
% demonstrate.

\bibliographystyle{IEEEbib}
\bibliography{refs}

\end{document}